\pdfoutput=1
\documentclass{article}

\PassOptionsToPackage{square,numbers}{natbib}

\usepackage[final]{neurips_2020_emecom}

\usepackage[utf8]{inputenc} 
\usepackage[T1]{fontenc}    
\usepackage{hyperref}       
\usepackage{url}            
\usepackage{booktabs}       
\usepackage{amsfonts}       
\usepackage{nicefrac}       
\usepackage{microtype}      

\usepackage{graphicx}
\usepackage{caption}
\usepackage{float}
\usepackage{subcaption}

\title{On (Emergent) Systematic Generalisation and Compositionality in Visual Referential Games with Straight-Through Gumbel-Softmax Estimator}

\author{%
  Kevin Denamganaï and James Alfred Walker \\
  Department of Computer Science\\
  University of York\\
  York, UK \\
  \texttt{kyd500@york.ac.uk}, \texttt{james.walker@york.ac.uk} \\
  
}

\begin{document}

\maketitle

\begin{abstract}
    The drivers of compositionality in artificial languages that emerge when two (or more) agents play a non-visual referential game has been previously investigated using approaches based on the REINFORCE algorithm and the (Neural) Iterated Learning Model. Following the more recent introduction of the \textit{Straight-Through Gumbel-Softmax} (ST-GS) approach, this paper investigates to what extent the drivers of compositionality identified so far in the field apply in the ST-GS context and to what extent do they translate into (emergent) systematic generalisation abilities, when playing a visual referential game. Compositionality and the generalisation abilities of the emergent languages are assessed using topographic similarity and zero-shot compositional tests.
    
    Firstly, we provide evidence that the test-train split strategy significantly impacts the zero-shot compositional tests when dealing with visual stimuli, whilst it does not when dealing with symbolic ones. 
    
    Secondly, empirical evidence shows that using the ST-GS approach with small batch sizes and an overcomplete communication channel improves compositionality in the emerging languages. Nevertheless, while shown robust with symbolic stimuli, the effect of the batch size is not so clear-cut when dealing with visual stimuli. Our results also show that not all overcomplete communication channels are created equal. Indeed, while increasing the maximum sentence length is found to be beneficial to further both compositionality and generalisation abilities, increasing the vocabulary size is found detrimental.
    
    Finally, a lack of correlation between the language compositionality at training-time and the agents' generalisation abilities is observed in the context of discriminative referential games with visual stimuli. This is similar to previous observations in the field using the generative variant with symbolic stimuli.
\end{abstract}

\section{Introduction}

In recent years, research into language emergence and grounding have received increased attention. The former of which raises the question of how to make artificial languages emerge with similar properties to natural languages, or at least `natural-like' protolanguages, exhibiting compositionality as the primarily-targeted property\citep{Baroni2019, Guo2019, Li&Bowling2019, Ren2020}. Indeed, languages' compositionality has been shown to further the learnability of said languages \citep{kirby2002learning,
Smith2003, Brighton2002, Li&Bowling2019} and promises to increase the generalisation ability of the artificial agent that would be able to wield them. For instance, it has been found to be instrumental in producing learned representations that generalise, when measured in terms of the data-efficiency of subsequent transfer and/or curriculum learning \citep{Higgins2017SCAN, Mordatch2017, MoritzHermann2017, Jiang2019}. Nevertheless, the ability of neural networks to generalise in a systematic fashion has been called into question~\citep{Lake&Baroni2018, Loula2018, Liska2018, Bahdanau2019}, and investigated towards finding necessary conditions and/or paradigms that favour the emergence of systematicity~\citep{Hill2019, Slowik2020, Korrel2019, Lake2019, Russin2019}. In this paper, similarly to \citet{Hill2019} investigating natural language grounding in the context of embodied agents, we take a closer look at the conditions that further the emergence of compositionality in artificial languages in the context of referential games.

\textbf{Straight-Through Gumbel-Softmax and Visual Stimuli.} Although it has been shown that emerging languages are far from being `natural'-like \citep{Kottur2017,Chaabouni2019a,Chaabouni2019b}, there are some successful cases demonstrating the emergence of compositional languages and learned representations (e.g. ~\citet{Kottur2017, Lazaridou2018, Choi2018, Bogin2018, Guo2019, Korbak2019, Chaabouni2020}),
relative to a given standard of compositionality. This paper focuses exclusively on the \textit{Straight-Through Gumbel-Softmax} (ST-GS) approach proposed by \citet{Havrylov2017}, as it supposedly allows a richer signal towards solving the credit assignment problem that language emergence poses since the gradient can be backpropagated from the listener agent to the speaker agent, while, in comparison, it cannot be backpropagated when using more commonly adopted approaches based on REINFORCE-like algorithms~\citep{williams1992simple}. 

\textbf{Symbolic vs Visual Stimuli.} The main works in language emergence and grounding focus on symbolic/one-hot-encoded stimuli (e.g. \cite{Kottur2017, Chaabouni2020}) whereas the ultimate goal of the field is related to grounding in similar modalities enjoyed by human beings, and primarily sight. Therefore, in order to take one more step in this direction, in this paper, we investigate primarily visual/pixel-based stimuli as well as whether results found in the context of symbolic stimuli translates into the context of visual stimuli.


\textbf{Compositionality \& Generalisation.} As a concept, compositionality has been the focus of many definition attempts. For instance, it can be defined as ``the algebraic capacity to understand and produce novel combinations from known components''(\citet{Loula2018} referring to \citet{montague1970universal}) or as the property that sees ``the meaning of a complex expression [a]s a function of the meaning of its immediate syntactic parts and the way in which they are combined''~\citep{krifka2001compositionality}. Although difficult to define, the commmunity seem to agree on the fact that it would enable learning agents to exhibit systematic generalisation abilities (also referred to as combinatorial generalisation \citep{Battaglia2018}). 
Some of the ambiguities that come with those loose definitions start to be better understood and explained, as in the work of~\citet{Hupkes2019}. In this paper, we will refer to compositionality as ``the ability to entertain a given thought implies the ability to entertain thoughts with semantically related contents''\citep{Fodor&Pylyshyn1988}, and thus use it interchangeably with systematicity, following the classification made by \citet{Hupkes2019}.

Compositionality, as a property of languages, can be difficult to measure. Like ~\citet{Keresztury2020}, we acknowledge that it ought to be measured more quantitatively than doing qualitative eye-balling of the artificial agents' utterances. Therefore, this paper aims to provide a more complete picture of the phenomena involving the ST-GS approach using two metrics:  \textit{topographic similarity} ~\citep{Brighton&Kirby2006}, which is acknowledged by the research community as the main quantitative metric for compositionality~\citep{Lazaridou2018, Guo2019,Slowik2020, Chaabouni2020, Ren2020}, and \textit{zero-shot compositional tests}, in which a set of stimuli composed of specific attributes are held-out from the training set, while making sure that the agents are still familiarising themselves with the specific attributes in different contexts/combinations to the held-out ones. 
The use of these two metrics has been shown to be critical by the concurrent work of \citet{Chaabouni2020}, which shows that learning agents are able to generalise in a referential game in spite of their utterances not being compositional, when measured with \textit{topographic similarity}.

\textbf{Contributions.} Firstly, this paper questions whether the train-test split strategy matters when using \textit{zero-shot compositional tests}. Out of the two strategies tested (see Section~\ref{sec:train-test-split-details}), it was found that the choice significantly impacts the metric when dealing with visual stimuli, whilst it does not when dealing with stimuli that are symbolic/one-hot-encoded (see Section~\ref{sec:exp:train-test-split}). Appendix~\ref{sec:gen-difficulty} further investigates this impact using the lens of the concept of generalisation difficulty proposed by \citet{Chollet2019}. 

Secondly, following the definition of compositionality/systematicity and the use of both topographic similarity and zero-shot compositional test metrics, this paper provides a quantitative report on the extent with which the ST-GS relaxation is a viable approach to make compositional languages emerge in a (discriminative) referential game. We start by identifying the batch size hyperparameter as a potential driver of compositionality (see  Section~\ref{sec:exp:batch-size}). Whilst our experiments depict it as a robust driver of compositionality in the context of symbolic stimuli, its impact on compositionality is less significant when dealing with visual stimuli. Subsequently, we investigate the impact on compositionality of both the level of structure in the observed stimuli and the capacity of the communication channel (see Section~\ref{sec:exp:struct-capacity}). Our experiments show that, using the ST-GS approach, they both correlate positively with the agents generalisation abilities. More specifically, we found that increasing the capacity of the communication channel by means of increasing the maximum sentence length correlates positively with both the compositionality of the language and the generalisation abilities of the learning agents, but increasing it by means of increasing the vocabulary size is negatively correlated with our two metrics. Appendix ~\ref{sec:language-emergence} provides theoretical insights on these phenomenon by comparing the ST-GS approach to the (Neural) \textit{Iterated Learning Model} (ILM)~\citep{Kirby&Hurford2002,Ren2020}.

Finally, similarly to ~\citet{Hupkes2019}, this paper investigates under what circumstances a model can be called compositional, and more precisely when does a learning agent is said to have systematicity. 
We ask whether a learning agent has systematicity/compositional abilities when the ``utterances in the arena of use''~\citep{hurford1987language} that it produces are compositional? 
While there seems to be such an implicit assumption in the research community
, the results presented in Section~\ref{sec:exp:gen-compo-train} disagree with this assumption. This is similar to the results observed in the concurrent and related work of ~\citet{Chaabouni2020}, in which symbolic/one-hot-encoded stimuli and a generative referential game are used, instead of visual stimuli in a discriminative framing as in this paper (see ~\citet{Denamganai2020a} for further details on the different variants of referential games). 


\section{Preliminaries}
\subsection{Visual Referential Games}
\label{sec:visual-referential-games}

\begin{figure*}[t]
    \centering
    \begin{subfigure}[t]{0.7\linewidth}
        \includegraphics[width=1.0\linewidth]{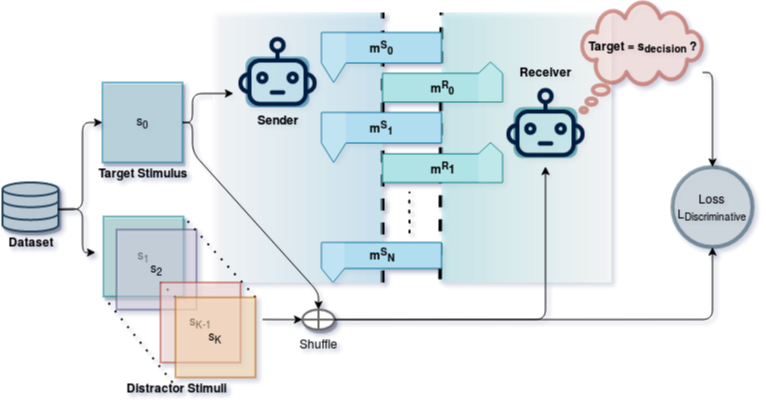}
    \end{subfigure}
    \begin{subfigure}[t]{0.29\linewidth}
        \vspace{-140pt}
        \includegraphics[width=1.0\linewidth]{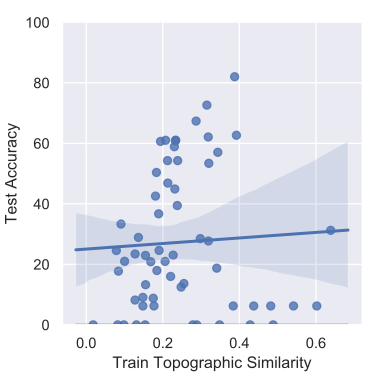}
    \end{subfigure}
    
    \begin{subfigure}[t]{0.49\linewidth}
        \includegraphics[width=0.49\linewidth]{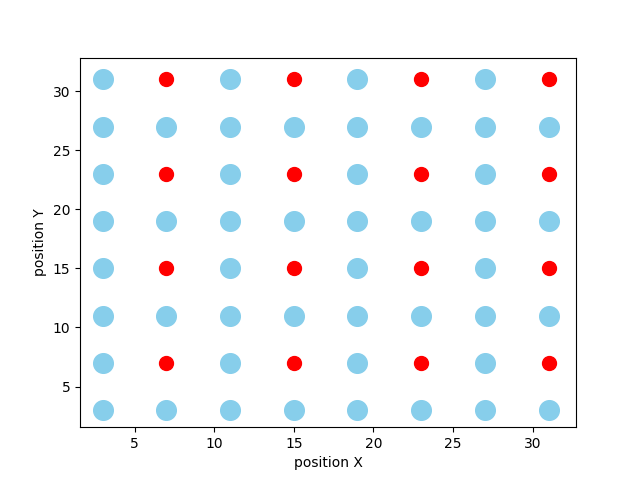}
        \includegraphics[width=0.49\linewidth]{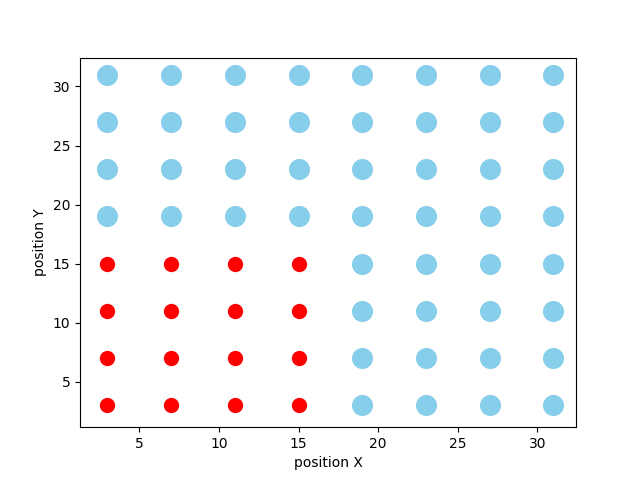}
    \end{subfigure}
    \begin{subfigure}[t]{0.49\linewidth}
        \includegraphics[width=0.49\linewidth]{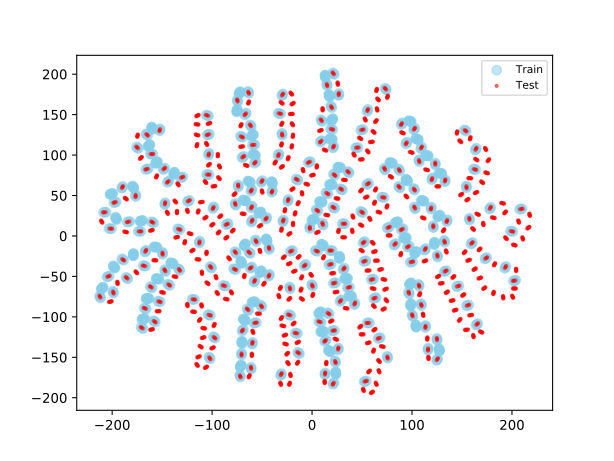}
        \includegraphics[width=0.49\linewidth]{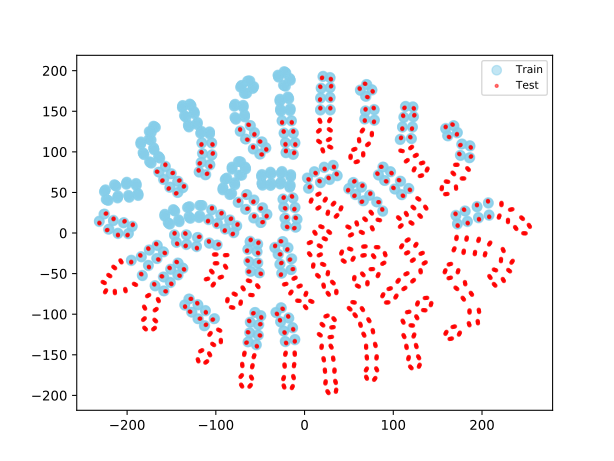}
    \end{subfigure}
    
    \caption{\textbf{Top left:} Illustration of a \textit{partially-observable $2$-players/$L$-signal/$N$-round/uniformly-distributed-distractors} \textit{referential game}. \textbf{Top right:} Test-time accuracy with respect to the training-time topographic similarity across the different settings, along with a linear regression model fit for comparison.
    \textbf{Bottom:} $2D$ projection of the symbolic $2$-attribute benchmarks (left) and T-SNE representations of the symbolic $3$-attribute benchmarks (right),  for both \textit{interpolation} and \textit{extrapolation} tasks (from left to right). Blue dots correspond to training examples (their vicinity is emphasised by drawing them larger in the t-SNE), while red dots correspond to testing examples.}
    \label{fig:extra-vs-inter}
    \label{fig:ReferentialGame}
    \label{fig:gen-f-ts}
\end{figure*}

Referential/Language games emphasise the functionality of languages, namely, the ability to efficiently communicate and coordinate between agents. 
Following the nomenclature proposed in \citet{Denamganai2020a}, we will focus primarily on a \textit{partially-observable $2$-players/$L$-signal/$0$-round/uniformly-distributed-distractors} variant illustrated in Figure~\ref{fig:ReferentialGame}.
As it is a discriminative referential game, we specify the number of distractor stimuli $K$. At training-time, $K=47$, and at testing-time, $K=63$, which corresponds to the maximum number of distractors that can be used across the different benchmarks, which are described in Section~\ref{sec:evaluation-methods}. For a fair comparison of the measure of accuracy between benchmarks, the values of $K$ are fixed. Details about the agent's architecture can be found in Appendix~\ref{app:model-architecture} and in our code\footnote{Our code is released at: \url{https://github.com/Near32/ReferentialGym/tree/master/zoo/referential-games\%2Bst-gs}}. Having described the setup, the following section provides details of the kind of stimuli and the train/test set split that is used to evaluate the generalisation abilities of the tested learning agents, whilst emphasising where this study lies in the spectrum of generalisation evaluation.


\subsection{Generalisation Abilities}
\label{sec:generalisation}

While many studies have investigated the generalisation abilities of neural networks in the context of linguistic tasks (e.g. ~\citet{Kottur2017, Lake&Baroni2018, Resnick2019}), their assumed definition to the act of generalisation do not always coincide~\cite{Hupkes2019}. Following the work of ~\citet{Lake&Baroni2018} showing that recurrent neural networks (RNNs) fail to generalize systematically but are very successful at generalizing when enough supporting evidence is provided. It is worth emphasizing again, like many previous authors ~\citep{Bahdanau2019, Hupkes2019, Russin2019, Chollet2019}, that it is critical to clearly state what kind of generalisation abilities this study aims to evaluate. The remainder of this section proposes an in-depth examination of the range of possible generalisation abilities and suggests where this paper lies within those identified.

While some studies test for what is referred to as systematic generalisation abilities ~\cite{Kottur2017, Lake&Baroni2018, Liska2018, Korrel2019, Bahdanau2019, Russin2019}, others seem to test for (vanilla) generalisation abilities~\cite{Andreas2019, Hupkes2019, Chaabouni2020}. This work differentiates between the two in terms of the size of the pool of supporting evidence/samples, which the learning system is trained on. Note that pool size is assumed to vary the difficulty of the task. Thus, (hard) systematic generalisation being on the side of the spectrum where the pool size is the smallest, and (easy) (vanilla) generalisation being on the other side where it is the greatest. In this work, we focus on systematic generalisation, rather than vanilla, and we adopt ~\citet{Bahdanau2019}'s definition of systematic generalisation abilities (i.e. a model that has systematic generalisation abilities  ``should be able to reason about all possible object combinations despite being trained on a very small subset of them''). This definition is a continuation of our assumed definition for compositional abilities, particularly ``the ability to entertain a given thought implies the ability to entertain thoughts with semantically related contents''\citep{Fodor&Pylyshyn1988}. In other words, the kind of generalisation abilities we propose to test for could be referred as systematic combinatorial/compositional generalisation abilities.




\subsection{Emergent Systematicity \& Prior Knowledge}
\label{sec:gen:emergent-systematicity}

The work of ~\citet{Hill2019} is particularly relevant to our current context, as it evaluates a specific kind of systematic generalisation referred to as ``emergent systematicity'' (~\citet{Hill2019} citing ~\citet{mcclelland2010letting}), ``because the architecture of our agent does not include components that are explicitly engineered to promote systematicity'' (i.e. systematic generalisation abilities). The concurrent work of ~\citet{Chaabouni2020} also tackles ``emergent'' systematicity/systematic generalisation. The learning agents will be equipped with CNNs and RNNs (see details in Appendix~\ref{app:model-architecture}), giving them some form of spatial/translational invariance prior, hierarchical/distributed visual feature prior, and sequence-computation-enabling prior. 

\citet{Chollet2019} emphasizes that \textbf{priors}, \textbf{experience}, and \textbf{generalisation difficulty} must be controlled in order to reliably evaluate broad generalisation abilities. Acknowledging the context of emergent systematicity in which this study lies can therefore be understood as a step towards controlling the prior knowledge baked into deep learning agents when evaluating their generalisation abilities. While the matter of controlling the generalisation difficulty is difficult to address in the current context, it is broached upon in Appendix~\ref{sec:gen-difficulty}. The matter of how to control each for the amount of experience available to the agent is discussed in Section~\ref{sec:evaluation-methods}.

\section{Evaluation Methodology}
\label{sec:evaluation-methods}

In the following experiments, learning agents will observe both symbolic and visual stimuli from particular train/test splits of the dSprites dataset~\citep{dsprites17,Higgins2016}. Symbolic representations are obtained by one-hot-encoding the latent representations that the dataset provides. Originally employed as a benchmark for disentangled representation learning, the dSprites dataset consists of visual representations for combinations of values along some generative factors/attributes/latent axes. There are $32$ possible values on each position axis, X and Y, $40$ possible values on the Orientation axis, $6$ possible values on the Scale axis, and $3$ possible values on the Shape axis. Note that in our experiments the value on the Shape attribute is always fixed to be the \textit{heart} shape in order to remove any orientation ambiguities (e.g. the square shape has four symmetries that prevents visual differentiation from, for instance, a rotation of $90$ degrees and that of $180$ degrees). The choice of the dSprites dataset is motivated by the availability of the generative factors (that are used as symbolic stimuli in their one-hot-encoded form), which permits the computation of topographic similarity~\citep{Brighton&Kirby2006} (following a similar approach to \citet{Lazaridou2018}) to assess the degree of compositionality in the emerging languages, as well as doing so separately from assessing generalisation abilities. We emphasise again how critical the separation of the two measures is given the related and concurrent work of \citet{Chaabouni2020} showing that learning agents are able to generalise in a referential game in spite of their utterances not being compositional.




The amount of prior experience will be controlled in two ways. Firstly, by limiting the diversity in the training set, thus constraining our study to very small data set sizes. Secondly, by limiting the number of training epochs to fit to a sample budget of $480,000$ training samples, across the different settings. The number of gradient steps is the sample budget divided by the batch size assumed, which is detailed for each experiments in Section~\ref{sec:exp}.




\subsection{Train-Test Split Strategy}
\label{sec:train-test-split-details}

In the first experiment, we question whether the train-test split strategy matters when testing zero-shot compositional abilities. We arbitrarily propose to look at two simple train-test split strategies leading up to two different sets of tasks that we will refer to as: the interpolation tasks, $T_{inter}$, and the extrapolation tasks, $T_{extra}$. 
Considering that each stimulus can be described as a set of values taken from each attribute/generative factor/latent axis of the dSprites dataset, details of how the train-test splits are performed in each task can be described as follows. In the interpolation tasks, $T_{inter}$, the test sets consist of defining testing-purpose (and not test-only) values alternating with other values, on each attribute axis. In the extrapolation tasks, $T_{extra}$, the test sets consist of defining the testing-purpose (and not test-only) values as the first values (following the ordering of the original dataset) on each attribute axis. We highlight already, and will detail further below, that testing-purpose values are encountered by the agent both at training-time and testing-time, just not in combinations with all other possible values on the other latent axises of which a stimuli is composed.

Independently of the task, half of the possible values for each attribute axis are defined as testing-purpose values. The choice of using half of the possible values is motivated by the results of \citet{Bahdanau2019} when evaluating for emergent systematicity, i.e. a CNN+LSTM architecture similar to that of ours here, and showing that such a benchmark is already challenging enough.
Then, any sampled stimulus that combines \textbf{two or more} testing-purpose values for its different attributes is automatically retained for the test set. In other words, the testing-purpose values on each latent axis are presented to the agent at training-time while in combinations with sole non-testing-purpose values on the other latent axises, in order for the agent to familiarise itself with all the possible values on each latent axis while remaining unaware of a subset of all the combinations possible.  For an intuitive understanding, Figure~\ref{fig:extra-vs-inter} renders $2D$ projection of the symbolic $2$-attributes tasks, $T^{2}_{inter}$ and $T^{2}_{extra}$, and the results of performing t-SNE~\citep{maaten2008visualizing} on the symbolic $3$-attributes tasks, $T^{3}_{inter}$ and $T^{3}_{extra}$, thus highlighting their systematic and topological differences, and revealing the motivation for their naming as interpolation and extrapolation tasks. 
In a similar manner to \citet{Russin2019}, we acknowledge the intuitive difference of difficulty between the two flavours in which generalisation occurs~\cite{marcus2018deep}. These are:  \textit{interpolation}, where ``the train and test sets are independent and identically distributed (i.i.d)'', and \textit{extrapolation}, where the tested system is required to make ``an inferential leap about the entire structure of part of the distribution that they have not seen'', and so the test set is out-of-domain (o.o.d.) with regards to the train set. When testing for combinatorial generalisation abilities, following \citet{Chollet2019}'s framework, it is hypothesised (and further detail in Appendix~\ref{sec:gen-difficulty}) that tasks involving compositional generalisation by \textit{interpolation}, $T_{inter}$, have \textit{zero generalisation difficulty}, while tasks involving compositional generalisation by \textit{extrapolation}, $T_{extra}$, bear \textit{some generalisation difficulty}. We emphasise again that the naming convention adopted here for the tasks is solely motivated by the intuition provided by figure~\ref{fig:extra-vs-inter}, but the actual nature of the tasks are unknown, since we are dealing with stimuli represented as combinations of attributes. They cannot and should not be confounded with regular interpolation and extrapolation tasks, as far as we know.

\subsection{Control for the Level of Structure in the Meaning Space}

In the subsequent experiments, we will vary the level of structure in the meaning space by experimenting with subsets of the dSprites dataset, emphasizing either: $2$ attributes (position on the X-axis and Y-axis, yielding $48$ training examples and $16$ testing examples), $3$ attributes (similar to $2$ attributes plus Orientation, yielding $256$ examples in each train/test set), or $4$ attributes (similar to $3$ attributes plus Scale, yielding $960$ training examples and $2112$ testing examples). We leave it to future works to vary the level of structure by varying the number of possible values for each attribute.

The employed subsets subsample the original dSprites dataset, in order to reduce the number of samples available to the learning agents. This subsampling aims to control for the amount of prior experience, and thus design a test where the learning agents are not presented with oversized data sets from which to slowly build evidence ~\citep{Lake&Baroni2018, Loula2018, Liska2018}. With the exception of the Scale attribute that only originally contains $6$ possible values (and thus remains unchanged), each other attribute contains $8$ possible values that are sampled from the original data to be evenly spaced out (i.e. out of the $32$ possibles values on the X and Y attributes, we sample every $4$ values; out of the $40$ possibles values on the Orientation attribute, we sample every $5$ values). 

\subsection{Control for the Capacity of the Communication Channel}

Two different cases are considered in the following experiments. The \textit{complete} case, where there are exactly $8$ ungrounded symbols in the vocabulary $V$, plus a ninth grounded symbol, in order to account for the \textit{end of sentence} semantic, thus $|V|=9$. The maximum sentence length $L$ is always equal to the number of attributes in the subset on which the experiment takes place. On the other hand, the \textit{overcomplete} case consists of a vocabulary of size $|V|=100$, and maximum sentence length $L=20$.

\begin{figure}[t]
    \centering
    \begin{subfigure}[t]{0.24\linewidth}
        \includegraphics[width=1.0\linewidth]{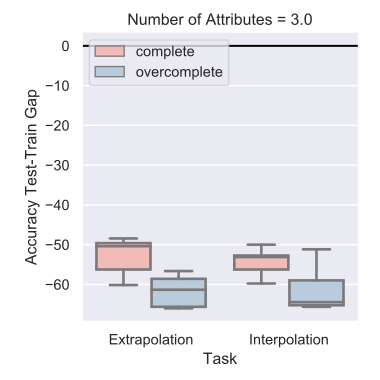}
    \end{subfigure}
    \begin{subfigure}[t]{0.24\linewidth}
        \includegraphics[width=1.0\linewidth]{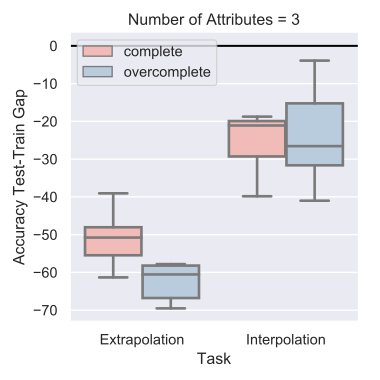}
    \end{subfigure}
    \begin{subfigure}[t]{0.24\linewidth}
        \includegraphics[width=1.0\linewidth]{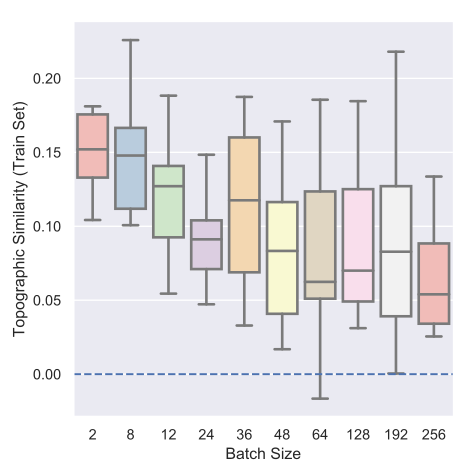}
    \end{subfigure}
    \begin{subfigure}[t]{0.24\linewidth}
        \includegraphics[width=1.1\linewidth]{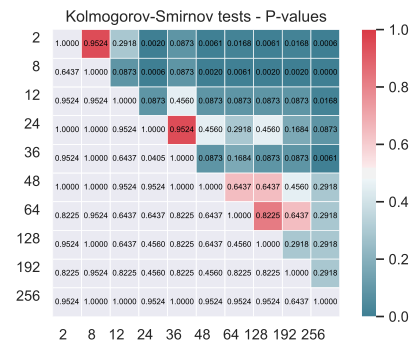}
    \end{subfigure}
    
    \begin{subfigure}[t]{0.24\linewidth}
        \includegraphics[width=1.0\linewidth]{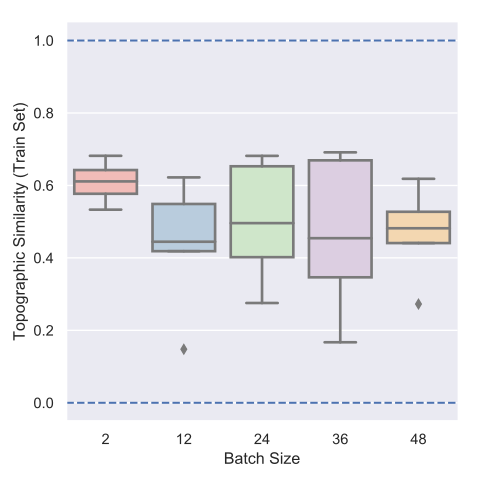}
    \end{subfigure}
    \begin{subfigure}[t]{0.24\linewidth}
        \includegraphics[width=1.05\linewidth]{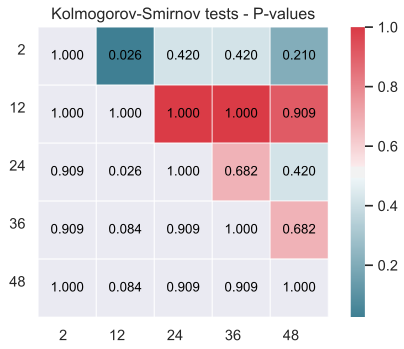}
    \end{subfigure}
    \begin{subfigure}[t]{0.24\linewidth}
        \includegraphics[width=1.0\linewidth]{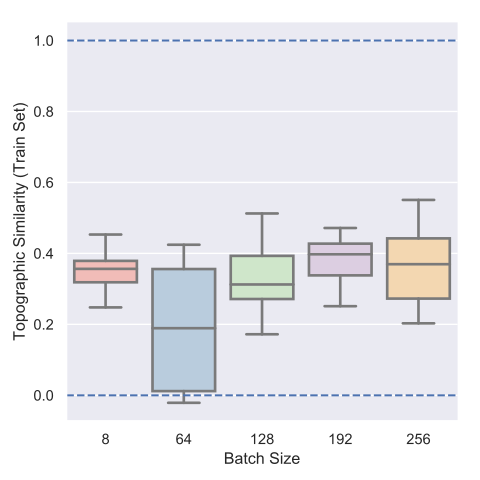}
    \end{subfigure}
    \begin{subfigure}[t]{0.24\linewidth}
        \includegraphics[width=1.1\linewidth]{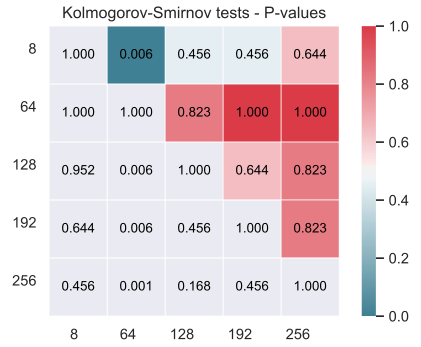}
    \end{subfigure}
    
    \caption{\textbf{Top left}: Distributions of the test-train gaps in accuracy measured on the 3-attributes benchmarks with, ordered from left to right, symbolic stimuli and visual stimuli.
    \textbf{Top right:} Distributions of training-time topographic similarity measured on the 3-attributes benchmarks with symbolic stimuli.
    The heatmaps illustrates the p-values of the KS two-samples tests between batch sizes with the alternative hypothesis being that the row-id distribution is `greater' than the column-id one.
    %
    \textbf{Bottom}: Distributions of training-time topographic similarity measured, with respect to different batch sizes and levels of structure, on the 2- (\textbf{left}) and 3-attributes (\textbf{right}) benchmarks. The heatmaps illustrate the p-values of the KS two-samples tests between batch sizes with the alternative hypothesis being that the row-id distribution is `greater' than the column-id one.
    }
    
    \label{fig:ttsplit-visual-attr3}
    \label{fig:ttsplit-symbolic-attr3}
    
    \label{fig:symbolic-attr3-batch-size-distr}
    \label{fig:symbolic-attr3-batch-size-KS}
    
    \label{fig:batch-size-attr2}
    \label{fig:batch-size-attr3}
\end{figure}

\section{Experiments}
\label{sec:exp}

\subsection{Experiment 1: Impact of Train-Test Split Strategy}
\label{sec:exp:train-test-split}

We firstly question whether the train-test split strategy matters when using \textit{zero-shot compositional tests}. 
We investigate in the contexts of $3$ attributes with symbolic and visual stimuli. Figure~\ref{fig:ttsplit-visual-attr3}(top-left) shows that, independently of the capacity of the communication channel, the choice of the train-test split significantly impacts the metric when dealing with visual stimuli, as a wide gap of performance can be seen between the two tasks, whilst it does not when dealing with symbolic stimuli.

\subsection{Experiment 2: Drivers of Compositionality}
\subsubsection{Effect of the Batch Size}
\label{sec:exp:batch-size}

\begin{figure}[t]
    \centering
    \begin{subfigure}[t]{0.49\linewidth}
        \hspace{-20pt}
        \includegraphics[width=1.0\linewidth]{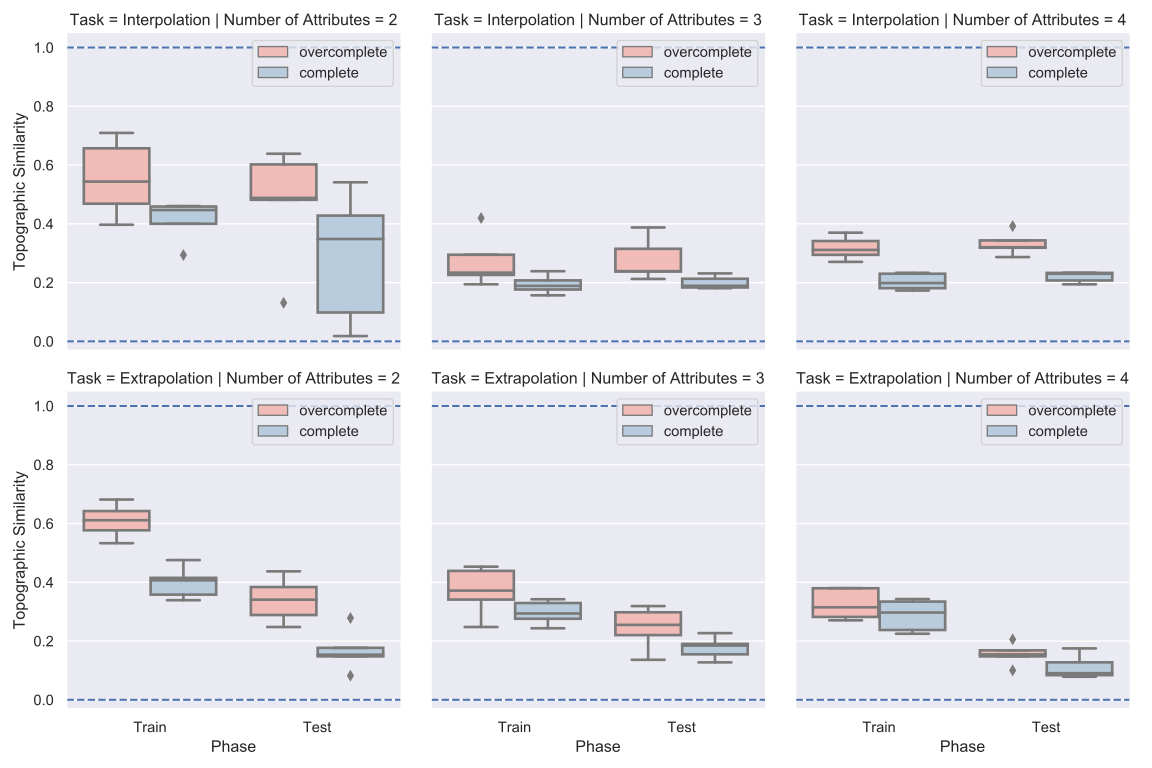}
    \end{subfigure}
    \begin{subfigure}[t]{0.49\linewidth}
        \centering
        \includegraphics[width=1.0\linewidth]{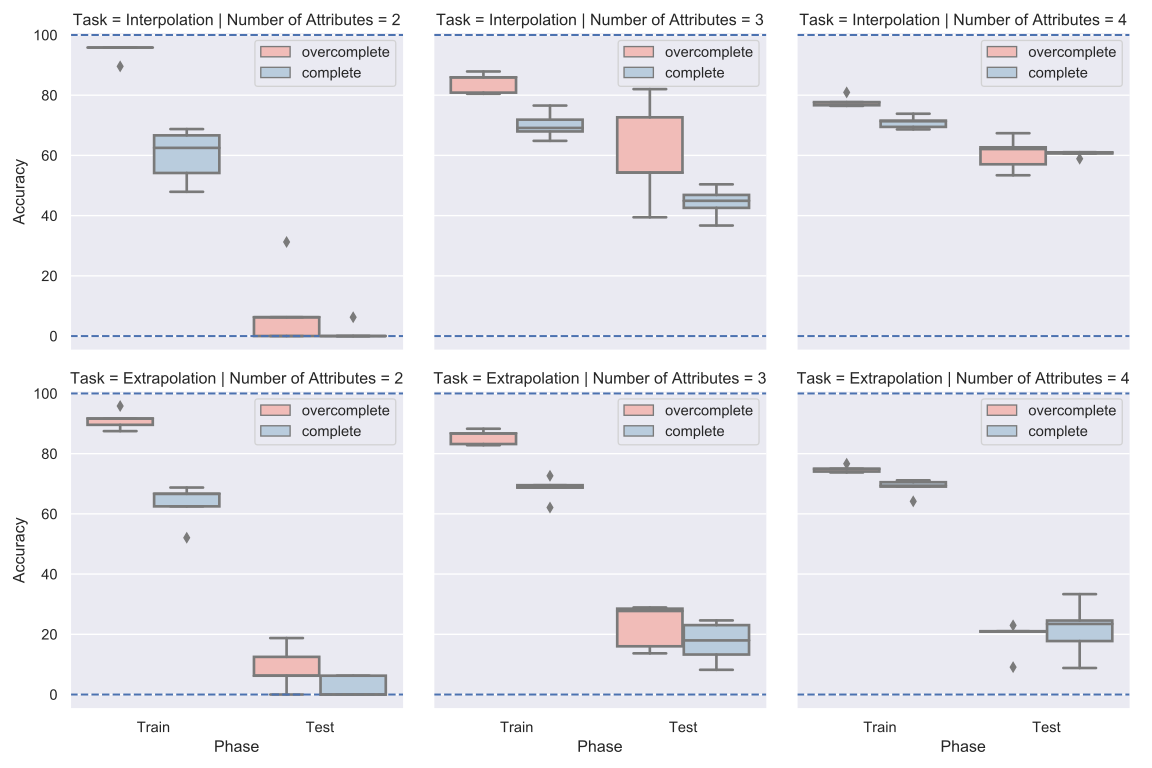}
    \end{subfigure}
    \caption{Distributions of the topographic similarity (left) and accuracy (right), across the different benchmarks, and the different communication channels.}
    \label{fig:toposim-struct-cap}
    \label{fig:acc-struct-cap}
\end{figure}

In this section, we investigate the hypothesis that the batch size modulates a transmission bottleneck effect in the context of the ST-GS algorithm, similarly to the effect highlighted in the ILM where the severity of the bottleneck (or its inverse, the coverage) is positively (negatively) correlated with the likelihood of highly-compositional language emergence. The soundness of this hypothesis is detailed in Appendix~\ref{subsec:transmission-bottleneck}. 
We firstly report on the the 3-attributes setting, with $10$ random seeds. Results obtained with symbolic and visual stimuli are presented, respectively, in figure~\ref{fig:symbolic-attr3-batch-size-distr} (right) and figure~\ref{fig:batch-size-attr3} (right). 
Once again, the dynamics are different, depending on the nature of the stimuli. While the symbolic context shows great support to the identification of the batch size as a robust driver of compositionality, its impact is less clear-cut when dealing with visual stimuli, as most of the p-values in our Kolmogorov-Smirnov (KS) two-samples tests are high. In this visual setting, 
we suspect that the batch normalization layers in the learning agents' CNNs starts playing a greater regularisation role as, in this 3-attributes setting, even the smallest coverage value already entails to relatively high batch sizes.

Knowing that the level of structure in the observed meaning space has also been highlighted as a driver of compositionality in the context of the ILM, we further investigate the context of visual stimuli by benchmarking against the 2-attributes setting, this time with $5$ random seeds for each batch size. Figure~\ref{fig:batch-size-attr2} (left) illustrates the results. 
Although not always statistically significant, in this level of structure, the likelihood of seeing highly-compositional languages seems to negatively correlate with the severity of the transmission bottleneck.

Comparing the results across the different levels of structure, it is surprising to see that, the topographic similarity measured is significantly lower when the level of structure is higher ($p\approx3\times10^{-4}$ on a KS test between benchmarks, at the lowest $\sim4\%$ coverage). Thus, on the contrary to the phenomenon observed in the ILM context, the ST-GS approach does not necessarily further higher compositionality in higher levels of structure. Finally, independently of the level of structure, the lowest coverage always yields relatively high topographic similarity languages, with relatively low variance in the distributions. Therefore, we subsequently fix the batch size to accommodate those lowest coverage values around $4\%$ (i.e. $2$ in the 2-attributes, $8$ in the 3-attributes, and $40$ in the 4-attributes benchmarks).

\subsubsection{Effect of Input/Meaning Space Structure and Communication Channel Capacity}
\label{sec:exp:struct-capacity}

We investigate the impact of both the level of structure in the observed meaning space and the capacity of the communication channel.
Figure~\ref{fig:toposim-struct-cap} (left) illustrates the impact in terms of topographic similarity. The most striking result is that, independently of the task, the overcomplete communication channel promotes significantly higher compositionality in the emerging languages ($p\approx9\times10^{-4}$ and $p\approx0.037$ for one-sided KS tests with the alternative hypothesis being that the overcomplete channel yields greater topographic similarity than the complete one, respectively, at testing-time and training-time, for the interpolation task ; $p\approx0.0137$ and $p\approx0.037$, similarly, for the extrapolation task). As the level of structure increases, the compositionality of the emerging languages using the ST-GS approach decreases, which is contrary to what can be observed in the context of the ILM. 

In terms of the generalisation abilities of the learning agents, we report the results in figure~\ref{fig:acc-struct-cap} (right). 
The ST-GS approach seems to significantly favour overcomplete communication channels over complete ones. Independently of the task, overcomplete communication channels significantly yield greater data-efficiency at training time ($p\approx1\times10^{-7}$ and $p\approx6\times10^{-9}$ for one-sided KS tests with the alternative hypothesis being that the overcomplete channel yields greater accuracy than the complete one, on the interpolation and extrapolation tasks, respectively). The effect appears to diminish as the level of structure in the observed meaning space increases. 
We observe a decrease of the test-train generalisation gap as the amount of structure in the meaning space increases. As this increase of structure implies an increase in size of the input/meaning space, our results provides evidence, this time in the context of visual stimuli, of the effect observed by \citet{Chaabouni2020} with symbolic stimuli that ``generalisation emerges `naturally' if the input space is large''\citep{Chaabouni2020}.

We further investigate the impact of varying the capacity of the communication channel when the input space structure is kept fixed, using the 3-attribute benchmarks, $T_{extra}^3$ and $T_{inter}^3$. In these experiments, we vary the maximum sentence length $L$, while keeping the vocabulary size $V$ fixed, and vice versa (see Appendix~\ref{sec:supplementary} for the graphs).
The results of conducting Spearman rank-order correlation tests with topographic similarity or with accuracy of zero-shot compositional tests are described in table~\ref{table:exp-comm-ch-spearman-rho-toposim}. Independently of the task, increasing $L$ translates mainly in an increase in topographic similarity, at training-time, as well as in an increase in systematicity, overall. On the other hand, increasing $V$ is marginally detrimental.


These results are all the more surprising given that, in the context of symbolic stimuli, REINFORCE-like algorithms have been shown to require constrained communication channels to yield highly compositional languages~\citep{Kottur2017}, and that its capacity negatively correlates with compositionality~\cite{Chaabouni2020}.
Here, using the ST-GS approach with visual stimuli, the emergence of both compositionality and systematicity is furthered by overcomplete channels with high maximum sentence length.

\begin{table}[t]
    \centering
    \caption{Results of Spearman rank-order correlation tests between the measured topographic similarity of the emerging languages (\textbf{top}), or the accuracy on the zero-shot compositional tests (\textbf{bottom}), and, respectively, the maximum sentence length $L\in\{3,6,9,20\}$ (for $V\in\{9,20\}$), or the vocabulary size $V\in\{9,20,50,100\}$ (for $L\in\{10,20\}$). $5$ seeds in each communication channel capacity.}
    \label{table:exp-comm-ch-spearman-rho-toposim}
    \label{table:exp-comm-ch-spearman-rho-acc}
    \begin{tabular}{@{}l r c c c c@{}} 
         \toprule
         Task & Phase & $V=9$ & $V=20$ & $L=10$ & $L=20$\\ 
         \midrule
         {$T^3_{inter}$} & Train    & $0.65$ ($p\approx 0.002$)     & $0.46$ ($p\approx 0.043$) 
                                    & $-0.18$ ($p\approx 0.452$)    & $-0.35$ ($p\approx 0.132$) \\
         \cline{2-6} & Test & $0.52$ ($p\approx 0.019$)     & $0.09$ ($p\approx 0.721$)
                            & $-0.07$ ($p\approx 0.770$)    &  $-0.21$ ($p\approx 0.376$) \\ 
         \hline
         {{$T^3_{extra}$}} & Train  & $0.63$ ($p\approx 0.003$)     & $0.50$ ($p\approx 0.023$) 
                                    & $-0.64$ ($p\approx 0.003$)    & $0.0$ ($p\approx 1.0$) \\ 
         \cline{2-6} & Test & $0.31$ ($p\approx 0.183$)     & $0.47$ ($p\approx 0.039$) 
                            & $-0.47$ ($p\approx 0.039$)    & $0.13$ ($p\approx 0.580$) \\ 
         \bottomrule
         \toprule
         \midrule
         {$T^3_{inter}$} & Train    & $0.87$ ($p\approx 10^{-6}$) & $0.73$ ($p\approx 10^{-4}$) 
                                    & $-0.02$ ($p\approx 0.948$)  & $-0.56$ ($p\approx 0.010$) \\
         \cline{2-6} & Test & $0.77$ ($p\approx 10^{-4}$) & $0.87$ ($p\approx 10^{-6}$) 
                            & $0.51$ ($p\approx 0.021$)   &  $-0.13$ ($p\approx 0.579$)\\ 
         \hline
         {{$T^3_{extra}$}} & Train  & $0.76$ ($p\approx 10^{-4}$) & $0.49$ ($p\approx 0.029$) 
                                    & $-0.43$ ($p\approx 0.060$)& $-0.05$ ($p\approx 0.845$)\\ 
         \cline{2-6} & Test & $-0.23$ ($p\approx 0.323$)  & $0.26$ ($p\approx 0.261$) 
                            & $0.18$ ($p\approx 0.442$)   & $0.13$ ($p\approx 0.590$) \\ 
         \bottomrule
        \end{tabular}
\end{table}

\subsubsection{Lack of Correlation between Generalisation and Compositionality at Training-time}
\label{sec:exp:gen-compo-train}


We now investigate the validity of the common assumptions that observing compositionality at training-time, as reported by topographic similarity, correlates with the systematicity of the learning agents. Figure~\ref{fig:gen-f-ts} (right) shows the test-time accuracy with respect to the training-time topographic similarity across the different settings (see Appendix~\ref{app:gen-f-ts-settings} for a break down by setting). The results of a Spearman-rho test yields a correlation coefficient of $0.136$ and $p\approx0.302$. Not only the coefficient is arguing towards a lack of correlation, the large p-value also prevents us from rejecting the non-correlation hypothesis, thus showing that it is incorrect to expect the training-time topographic similarity to predict any systematicity of the learning agents in the context of visual stimuli.



\section{Conclusion}

In this paper, we have shown that the choice of train-test split in zero-shot compositional tests significantly impacts the metric when dealing with visual stimuli, whilst it does not when dealing with symbolic ones. 
Then, we provided a quantitative report on the extent with which the ST-GS relaxation is a viable approach to make compositional languages emerge in a (discriminative) referential game. Fistly, while the batch size hyperparameter is identified as a critical driver of compositionality, its impact is significant in the context of symbolic stimuli but less so clearly understood when dealing with visual stimuli. Secondly, overcomplete communication channels with large maximum sentence lengths are found to be beneficial to further both compositionality and systematicity while those with large vocabulary sizes is found detrimental.
Our paper has highlighted many ways in which results about language emergence and grounding in the case of symbolic stimuli do not translate equally in the context of visual stimuli, thus raising the need for subsequent work in the visual context as vision is one of the main modalities of concerns for the grounding of languages that would be spoken by artificial agents living alongside and cooperating with human beings.

\section*{Broader Impact}

This work consists solely of simulations, thus evacuating some of the ethical concerns, as well as the concerns with regards to the consequences of failure of the system presented. With regards to the ethical aspects related to its inclusion in the field of Artificial Intelligence, we argue that our work aims to have positive outcomes on the development of human-machine interfaces, albeit being not yet mature enough to aim for this goal. The current state of our work does not allow us to extrapolate towards negative outcomes.

This work should benefit the research community of language emergence and grounding, in its current state.

\begin{ack}
This work was supported by the EPSRC Centre for Doctoral Training in Intelligent Games \& Games Intelligence (IGGI) [EP/L015846/1]. 

We would like to thank Dr Sondess Missaoui and the anonymous reviewers for their very helpful and constructive feedback on the draft of this paper.

We gratefully acknowledge the use of Python\cite{python-2009}, IPython\cite{ipython-perez-2007}, SciPy\cite{SciPy-NMeth2020}, Scikit-learn\cite{Scikit-learn:JMLR:v12:pedregosa11a}, Scikit-image\cite{scikit-image-van2014}, NumPy\cite{NumPy-Array2020}, Pandas\cite{pandas1-mckinney-proc-scipy-2010,pandas2-reback2020}, OpenCV\cite{opencv_library}, PyTorch\cite{pytorch-paszke-NEURIPS2019_9015}, TensorboardX\cite{huang2018tensorboardx}, Tensorboard from the Tensorflow ecosystem\cite{tensorflow2015-whitepaper}, without which this work would not be possible.
\end{ack}

\bibliography{bibli}

\appendix
\section{Supplementary Materials}
\label{sec:supplementary}

\begin{figure}[h]
    \centering
    \begin{subfigure}[t]{0.49\linewidth}
        \includegraphics[width=1.0\linewidth]{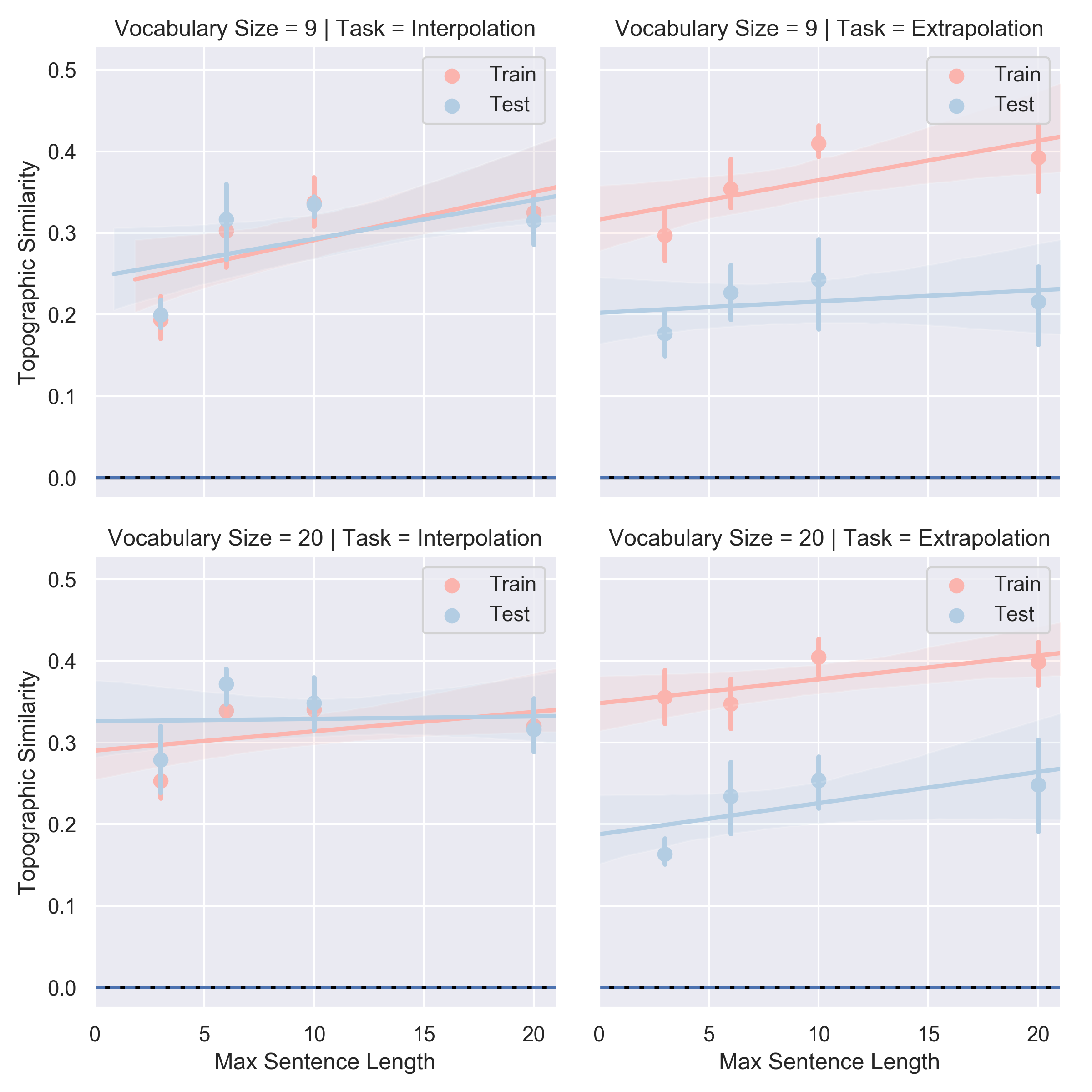}
    \end{subfigure}
    \begin{subfigure}[t]{0.49\linewidth}
        \centering
        \includegraphics[width=1.0\linewidth]{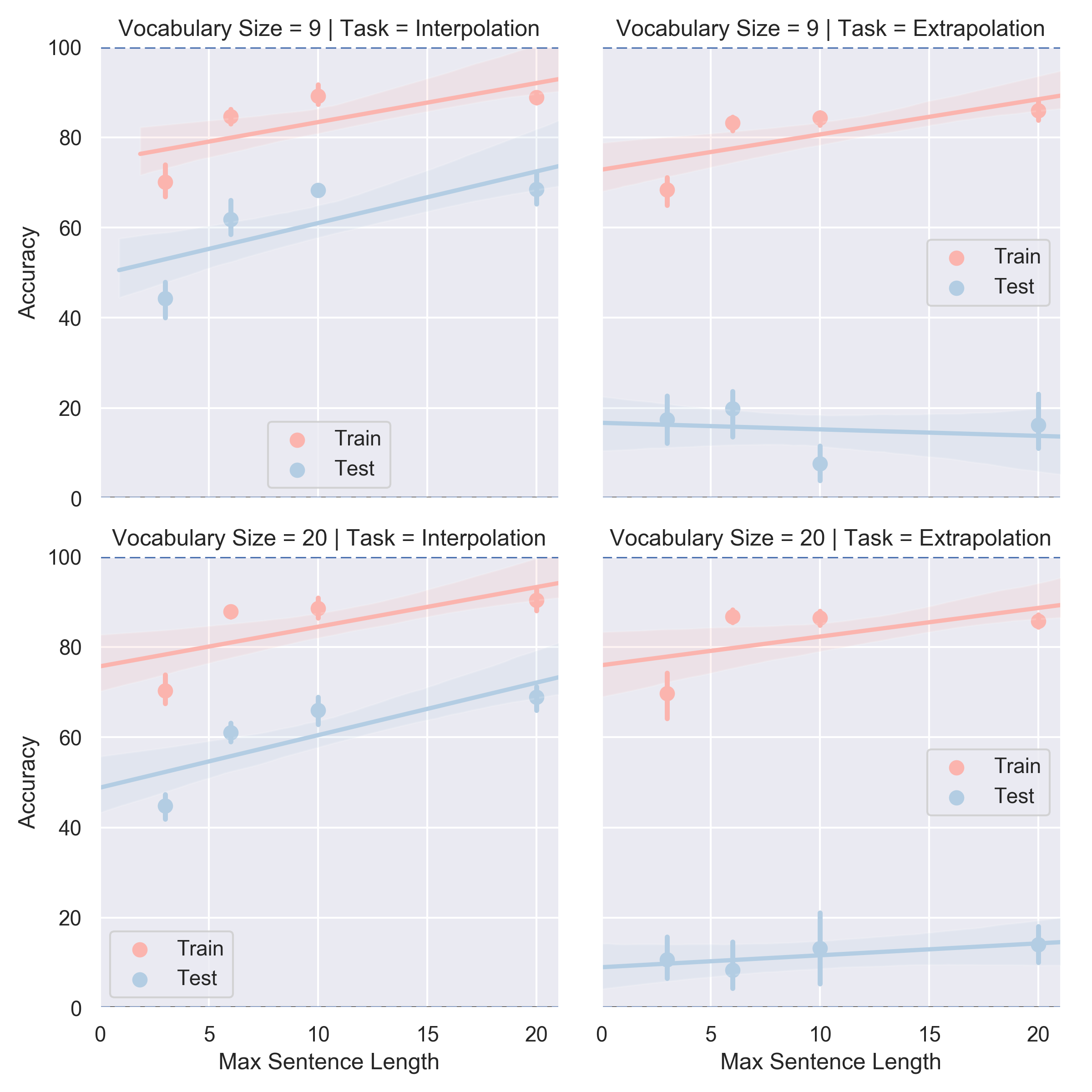}
    \end{subfigure}
    
    \begin{subfigure}[t]{0.49\linewidth}
        \includegraphics[width=1.0\linewidth]{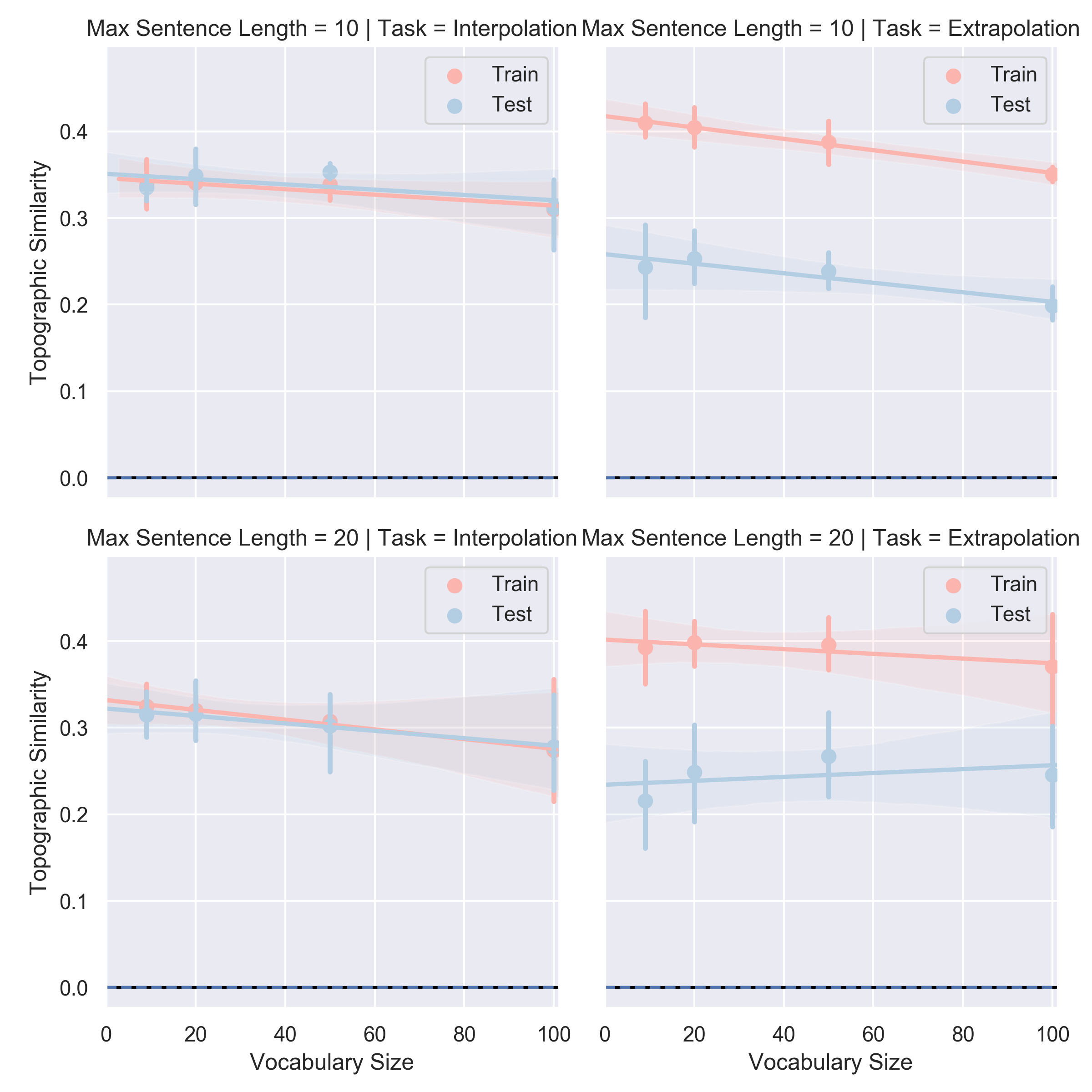}
    \end{subfigure}
    \begin{subfigure}[t]{0.49\linewidth}
        \centering
        \includegraphics[width=1.0\linewidth]{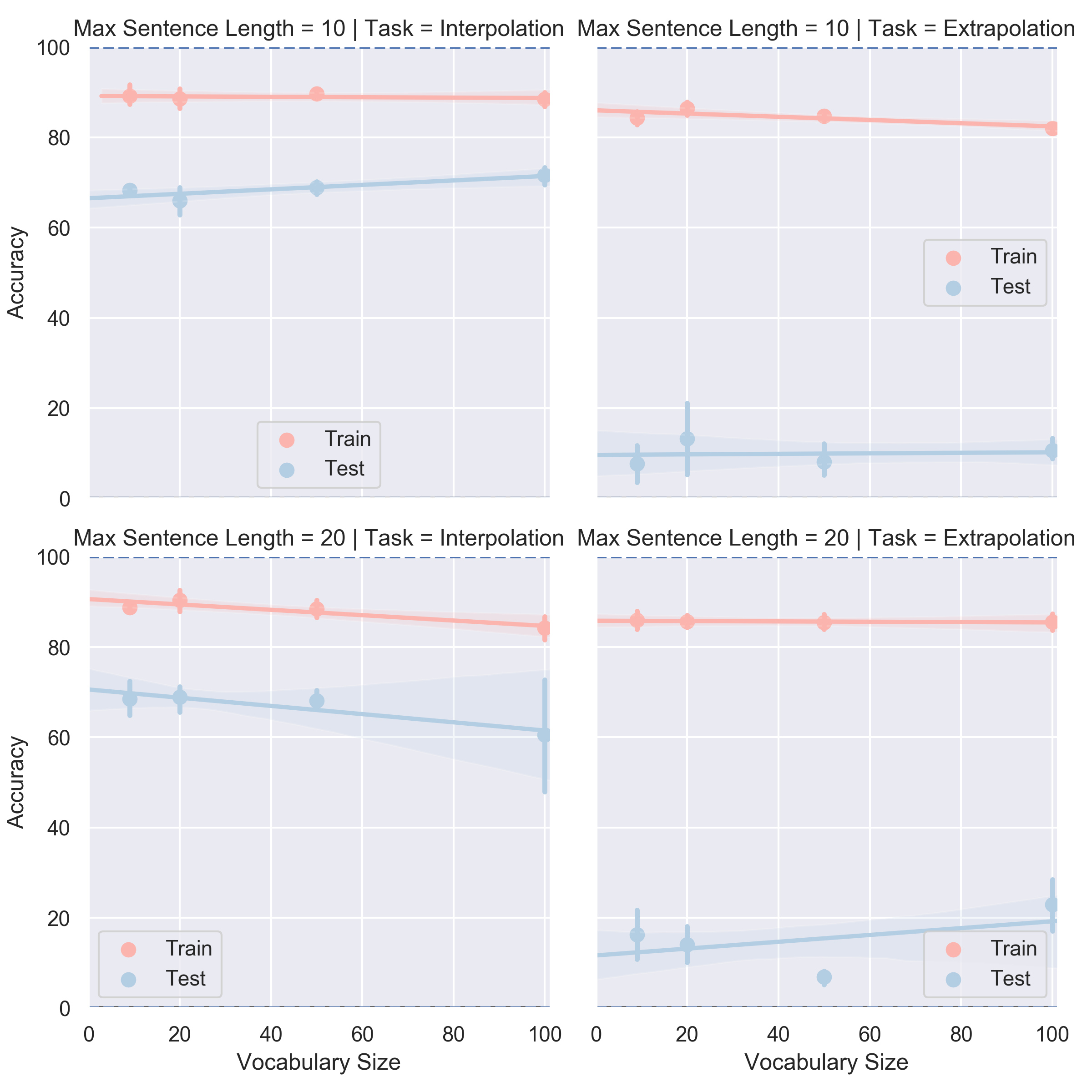}
    \end{subfigure}
    \caption{Distributions of the topographic similarity and accuracy, across different traversals of the communication channels capacities, on the 3-attributes benchmarks. \textbf{Top}: traversal of the max sentence length $L\in\{3,6,10,20\}$, for fixed $V\in\{9,20\}$. \textbf{Bottom}: traversal of the vocabulary size $V\in\{10,20,50,100\}$, for fixed $L\in\{10,20\}$.}
    \label{fig:exp-comm-ch-toposim}
    \label{fig:exp-comm-ch-acc}
\end{figure}

Figure~\ref{fig:exp-comm-ch-toposim} (top two rows) illustrates the impact on both the topographic similarity and the accuracy of the zero-shot compositional tests when varying the maximum sentence length $L$, while keeping the vocabulary size $V$ fixed, and vice versa (bottom two rows), as discussed in Section~\ref{sec:exp:struct-capacity}.

\section{Language Emergence}
\label{sec:language-emergence}
Compositionality and recursive syntax are the two main properties shared between all natural languages that account for their expressivity and flexibility. Among other things, they allow an ease of learning/acquisition as it can be seen in the \textit{poverty of the stimulus} phenomenon (``children master complex features of language on the basis of surprisingly little evidence" \citep{Brighton2002}).

When talking about language, it is common to differentiate between two language domains: I(Internal)-Language and E(External)-Language \citep{chomsky1986knowledge, hurford1987language, kirby2002learning} 
. I-Language is the internal representation a user has of the language it speaks and listens to, whilst E-Language is the set of external presentations/``utterances in the arena of use"\citep{hurford1987language} of language, when any of the users actually speak in an attempt to express a given meaning.

How language users are able to acquire an I-Language and are then able to contribute to the E-Language by producing utterances of language that exhibit compositionality is the question that the works of \citep{kirby2002learning, 
brighton2001survival,Brighton2002,Smith2003} strived to answer.  To achieve this they appealed to a ``process of information transmission via observational learning" \citep{Brighton2002}, which took place in-between populations and generations of (previously-learner) speaker agents and (soon-to-become-speaker) learner agents. This is the Iterated Learning Model (ILM)~\citep{Kirby2014}. 

In this section, we will compare the ST-GS estimator to the ILM in order to highlight insights about the drivers of compositional language emergence using the ST-GS estimator.

\subsection{(Neural) Iterated Learning Model}
\label{subsec:NILM}

The ILM consists of a ``[multi-]agent-based model where each agent represents a language user"\cite{Brighton2002}, which forms utterances based on the hypothesis that has been formed previously (I-Language) to account for previously-observed utterances (E-Language). It puts in relation an environment, made up of $N$ objects/stimuli that maps to a meaning space $\mathcal{M}=\{(f_i)_{i\in[1:F]} \; / \; \forall f_i, 1\leq f_i\leq V \}$, (consisting of an $F$-dimensional space, i.e. features, with $V$ possible discrete values along each dimension) with a signal space, or language, $\mathcal{S}=\{ (\omega_i)_{i\in [1:l]} \; / \; \forall\omega_i\in\Sigma \;,\; \forall l\in[1:l_{max}] \}$ (a set of strings of symbols from a symbol space $\Sigma$ with length $l\leq l_{max}$). The relation or mapping is induced by making agents observe (attend to) signal-meaning pairs from the E-Language, during the \textit{acquisition phase}, and then, during the \textit{production phase}, making them utter new language utterances, whilst being prompted to communicate/speak about a set of other randomly-sampled objects/stimuli from the environment. 

\subsubsection{Language Stability}

In such a context where language is evolved by each learner agent at each iteration, it becomes important to consider the stability of the different languages that may arise, in order to understand under which conditions does the ILM converge and what are the properties to expect of the emerging, stable languages. Before being able to consider the stability of a given language, we need to define its expressivity $E$, especially ``the expected number of meanings an individual will be able to express after observing some subset of the E-Language"\citep{Smith2003}, which we will denote $O$.

\textbf{Holistic Language} - In the context of a holistic language where there is no syntactic structure to the language and it consists of idiosyncrasies, no \textit{generalisation} can be operated, only \textit{memorisation} upon observation and then recall can help an agent express a given meaning in that language. Therefore, the expressivity of a holistic language $E_h$ ``is simply the probability of observing any particular meaning [$m\in O$, $Pr(m\in O)$,] multiplied by the number of possible meanings" \citep{Smith2003}, as shown in Eqn. \ref{eqn1}.
\begin{equation}
\label{eqn1}
    E_h = Pr(m\in O) \cdot {{V}\choose{1}}^F = Pr(m\in O) \cdot V^F.
\end{equation}
It is worth noting that the meaning space $\mathcal{M}$ is structured as an $F$-dimensional space of features with $V$ possible values for each feature.

\textbf{Compositional Language} - In the context of a compositional language where the relationship between meanings and signals is structured, ``expressivity becomes a function of the number of feature values observed, rather than a function of the number of meanings observed" \citep{Brighton2002}. Given a meaning $m=(v_1,\dots,v_F)$, formally, we have:
\begin{equation}
    E_c = Pr( ``\forall v\in m,\exists O_v\in R, v\in O_v") \cdot N_{used}
\end{equation}
where $Pr( ``\forall v\in m,\exists O_v\in R, v\in O_v") = Pr( ``\exists O_v \in R, v\in O_v" )^F$ is the probability of being able to express $m$ (to be contrasted with the probability of observing $m\in O$, $Pr(m \in O)$), and thus, $N_{used}$ is ``the expected number of expressible meanings" \citep{Brighton2002} that are used to label the $N$ objects in the environment. The more eager readers can refer to \citep{Brighton2002} for more details.

\subsubsection{Drivers of Compositionality} 

Since we are concerned with the likelihood of the emergence of compositionality over holisticity, the authors define in Eqn. \ref{eqn3} the relative stability $S$ of compositional languages with respect to holistic ones:
\begin{equation}
\label{eqn3}
    S = \frac{S_c}{S_c+S_h},
\end{equation}
where $S_c\propto \frac{E_c}{N}$ and $S_h \propto \frac{E_h}{N}$ are the (absolute) stabilities of compositional languages and holistic ones, respectively. Thus, ``the probability of observing some arbitrary meaning $m$, i.e. $Pr(m \in O)$, is determined by the number of objects in the environment ($N$), the number of meanings in the meaning space ($M$, where $M=V^F$), and the number of random object observations during an agent lifetime ($R$)" \citep{Brighton2002}. Therefore, the relative stability of a compositional language compared to a holistic language will be strongly affected by: (i) the structure of the meaning space, via its dependence to $M$, and (ii) the object coverage expressed by the ratio $b=\frac{R}{N}$ (also the inverse of the measure of the severity of the \textit{transmission bottleneck}). The lower the coverage, i.e. $b$, the greater the stability advantage of compositional languages over holistic ones is, such that ``the poverty-of-the-stimulus `problem' is in fact required for linguistic structure to emerge'' \citep{Smith2003}. Another important result is visible in the observation that ``a large stability advantage for a compositional language (high $S$) only occurs when the meaning space exhibits a certain degree of structure (i.e. when there are many features and/or values), suggesting that structure in the conceptual space of language learners is a requirement for the evolution of compositionality" \citep{Smith2003}.

In other words, both the severity of the (cultural) \textit{transmission bottleneck} and the degree of structure in the meaning space have been identified as drivers of compositionality in emerging languages, in the context of the ILM. 

Recently, the Neural Iterated Learning (NIL) algorithm was proposed by ~\citet{Ren2020}. They transposed the ILM framework to a deep learning setting and showed a similar impact of the severity of the transmission bottleneck, in addition to a proposed probabilistic framework that formally evidenced it.

\subsection{Straight-Through Gumbel-Softmax Estimator}
\label{sec:st-gs}

In the current computer science paradigm, we commonly assume a discrete nature of the messages that are sent by the \textit{speaker} to the \textit{listener}. Therefore, the communication channel is non-differentiable and must rely on \textit{Reinforcement Learning} algorithms to solve the credit-assignment problem. The most common candidates in the literature are REINFORCE-like algorithms \citep{williams1992simple}. Fortunately, tricks exist that allow the environment (in this case, the communication channel) to be made differentiable.

As it takes place in the context of an \textit{$L\geq2$-signal/$0$-round} referential game, the work of \citet{Havrylov2017} brings evidence that deep learning agents can not only learn to coordinate via a communication channel (as seen in \citet{Lazaridou2016} with ``atomic symbols'' \citep{Havrylov2017} already), but first and foremost invent a seemingly compositional, hierarchical, and variable-length language, with a ``sequence of tokens'' \citep{Havrylov2017}, in order to coordinate. The model's greatest success is attributed to the introduction of the Straight-Through Gumbel-Softmax (ST-GS) estimator/relaxation/backpropagation approach, which makes the communication channel differentiable (and also exhibits similar behaviour between training and testing, which is on the contrary to previous works, e.g.~\citet{Foerster2016}).

The ST-GS estimator is built on top of the Gumbel-Softmax estimator~\citep{jang2017categorical,maddison2017concrete}, that replaced one-hot-encoded symbols/tokens/words $w\in V$, originally sampled from a categorical distribution, with a continuous relaxation $\tilde{w}$, sampled from a Gumbel-Softmax distribution, following the notation of ~\citet{Havrylov2017}:
\[
    \tilde{w_k} = \frac{\exp((\log p_k + g_k)/\tau)}{\sum_{i=1}^{K}{\exp((\log p_i + g_i)/\tau)}}
\]
where $p_1,\dots,p_K$ are the $K$ event probabilities of the original categorical distribution, $\tau$ is a temperature hyperparameter (see ~\citep{Havrylov2017} for more details), and $g_1,\dots,g_K$ are sampled from the Gumbel distribution, i.e. $g_k=-\log(-\log(u_k))$ with $u_k\sim \mathcal{U}(0,1)$. 

Instead of stopping there, the ST-GS estimator performs a greedy discretization (i.e. using the $argmax$ operator) of $\tilde{w_k}$ during the forward pass, whilst relying on the continuous relaxation during the backward pass\footnote{for further implementation details see: \url{https://pytorch.org/docs/master/nn.functional.html?highlight=gumbel\#torch.nn.functional.gumbel_softmax}}, thus yielding a biased gradient estimator~\citep{jang2017categorical,bengio2013estimating}. 

\citet{Havrylov2017} simplified the whole process by considering learning the temperature $\tau(h_i^s)$ for each symbol/token/word $w_i$ in each sentence $s$ with conditioning on the hidden state $h_i^s$ of the sentence decoder RNN. In this work, instead of using a multi-layer perceptron, we only rely on a one-layer network $\alpha$: 
\[
    \tau(h_i^s) = \frac{1}{\tau_0+\log(1+\exp(\alpha(h_i^s)))}
\]

One very important and somewhat surprising result that they found is that using the ST-GS estimator, the greater the sequence length, the faster the learning of the communication protocol. On the other hand, there is no such correlation when training  REINFORCE-like algorithms. 
It is unclear whether this phenomena is due to the ST-GS estimator alone or the synergy between it and the pre-trained convolutional neural network (CNN), which each agent relied on in the original work. In order to efficiently assess the impact of the ST-GS estimator, in this paper we chose to have our agents learn everything from scratch.

\subsection{Instantiating a Transmission Bottleneck}
\label{subsec:transmission-bottleneck}

In the following, we highlight the similarities between the ILM, in the form of the NIL algorithm, and the ST-GS algorithm. Both the NIL and the ST-GS algorithms are iterative processes. Thus, we focus on one learning step for each.

Firstly, in the ILM/NIL, the \textit{learning phase} consists of the new learning agent $A_i$ updating its model using the data set $\mathcal{D}_{i-1} = (u^{i-1}_j, s_j)_{j\in [1;N_{learning}]}$ of pairs of stimuli $s$ and associated utterance $u$ produced by agent $A^{\prime}_{i-1}$ at the previous iteration.  $N_{learning}$ is the size of the learning data set, at each learning step, and it is randomly sampled from the whole meaning space $\mathcal{M}$ of size $N$. In the ILM/NIL, a cultural transmission bottleneck is instantiated by choosing $N_{learning}\leq N$, and the severity of the bottleneck is measured by $R=\frac{N_{learning}}{N}$. Next, following the naming of ~\citet{Ren2020}, in the \textit{transmitting phase}, the now updated agent $A^{\prime}_{i}$, fluent in the E-language $\mathcal{L}_{i-1}$ (as described by $\mathcal{D}_{i-1}$), is prompted to a production step where it generates the data set $\mathcal{D}_{i} = ( A^{\prime}_{i}(\hat{s}_j), \hat{s}_j)_{j\in [1;N_{learning}]}$ with stimuli being randomly sampled, $\hat{s}\sim\mathcal{M}$. Due to the transmission bottleneck during the production step, the learning agent has to generalise its knowledge of the E-language $\mathcal{L}_{i-1}$ to potentially novel stimuli, and it is thus bound to make this language evolve into a new E-language $\mathcal{L}_{i}$, unless there is no transmission bottleneck and it has perfectly learned the language during the learning step. In the NIL algorithm, an \textit{interacting phase} is intertwined between the \textit{learning phase} and the \textit{transmitting phase}, which consists of a referential game $\mathcal{R}$ that aims to promote disambiguation of the language. 

It is important to note that the ILM/NIL algorithm relies on the previous E-language $\mathcal{L}_{i-1}$ at each step $i$, whilst the ST-GS algorithm relies on something more akin to the previous I-language, as we will now detail.

In the context of the ST-GS algorithm, learning happens by batch. At each learning step, a batch $\mathcal{D}^{S}_{i}$ of stimuli, randomly sampled from the whole meaning space $\mathcal{M}$ (as defined in Eqn.~\ref{eq:learning-data-speaker}), is forwarded through the learning \textit{speaker} agent $S_{i}$ to produce a batch $\mathcal{D}^L_{i}$ of utterances (as defined in Eqn.~\ref{eq:learning-data-listener}). The latter is then forwarded through the learning \textit{listener} agent $L_{i}$, along with a batch of sets of stimuli $\Delta_i$. Each element of the batch is a shuffled list/set containing $K$ distractor stimuli that are randomly sampled from $\mathcal{M}$, and the corresponding target stimuli from $\mathcal{D}^S_i$ (as defined in Eqn.~\ref{eq:learning-data-delta}).
The resulting output is a batch of sets of scores $\Sigma_i$ (as defined in Eqn.~\ref{eq:scores}). Using $j$ as the index for the element of the batch, each score $s(u^i_j, \hat(s)^i_j(d)) \in \sigma^j_i$ intuitively represents the extent with which each utterance $u^i_j$ describes the stimuli $ \hat(s)^i_j(d) \in \delta^j_i$. 

\begin{equation}
    \label{eq:learning-data-speaker}
    \mathcal{D}^{S}_{i} = (s^i_j)_{j\in [1;N^S_{learning}]} \sim \mathcal{M}
\end{equation}
\begin{equation}
    \label{eq:learning-data-listener}
    \mathcal{D}^L_{i} = (u^i_j = S_{i}(s^i_j))_{j\in [1;N^S_{learning}]}
\end{equation}
\begin{equation}
    \label{eq:learning-data-delta}
    \Delta_i = (\delta^j_i)_{j\in[1;N^S_{learning}]} , s.t. \quad \forall j\in [1;N^S_{learning}], \delta^j_i = (\hat{s}^i_j(d))_{d\in [1;K+1]}
\end{equation}
\begin{equation}
    \label{eq:scores}
    \Sigma_i = (\sigma^j_i)_{j\in[1;N^S_{learning}]}
\end{equation}

Following the computation of the loss function, similarly to ~\citet{Havrylov2017}, the gradients are backpropagated and the agents are updated.
It is important to notice that at the time of the backward pass, the current agents are not yet updated, i.e. $L_{i}=L_{i-1}$ and $S_{i}=S_{i-1}$. Therefore, each of them receives a gradient with respect to the other's I-language (as represented by each agent's weights) at the previous time step, i.e. with respect to $S(\mathcal{L}_{i-1})$ and $L(\mathcal{L}_{i-1})$ respectively. Following the backward pass, the algorithm yields the updated agents $S^{\prime}_{i}$ and $L^{\prime}_{i}$.

In comparison to the NIL algorithm, we argue that both updated agents have received feedback in a richer fashion due to: (i) learning in a supervised fashion (that not only promotes but also penalises the E-language with respect to the goal of disambiguation) and (ii) the fact that the ST-GS relaxation enables feedback with respect to both the referential game $\mathcal{R}$ and the learning \textit{listener} agent's I-language $L(\mathcal{L}_{i-1})$ (which we assume to be more pertinent than the E-language $\mathcal{L}_{i-1}$).

\section{Agent Architecture}
\label{app:model-architecture}

Each agent consist of a language module and a visual or symbolic module, depending whether they deal with visual or symbolic stimuli. The \textit{listener} agent also incorporates a third decision module that combines the outputs of the other two (visual and language) modules. While the \textit{speaker} agent is prompted to produce the output string of symbols with a \textit{Start-of-Sentence} symbol and the visual module output as an initial hidden state, the \textit{listener} agent consumes the string of symbols with the null vector as the initial hidden state. In the following subsections, we detail each module architecture in depth.

\subsection{Visual \& Modules}

The visual module $f(\cdot)$ consists of four $3\times3$ convolutional layers with stride $2$. The two first layers have $32$ filters, whilst the last two layers have $64$. Each convolutional layer is followed by a $2$D batch normalisation layer, and the outputs are passed through a leaky ReLU activation function. Inputs are resized to $32\times32$, thus yielding feature maps of dimension $64\times2\times2$. The visual module outputs a flattened representation of dimension $256$. In a similar fashion, the symbolic module consists of a linear layer containing $256$ units followed by a leaky ReLU activation function.

\subsection{Language Module}

The language module $g(\cdot)$ consists of a one-layer LSTM network~\citep{hochreiter1997long} with $256$ hidden units, matching the dimension of the visual module output. 
In the context of the \textit{listener} agent, the input message $m=(m_i)_{i\in[1,L]}$ (produced by the \textit{speaker} agent) is represented as a string of one-hot encoded vectors of dimension $|V|$ and embedded in an embedding space of dimension $256$ via a linear layer and dropout layer of probability $p=0.8$. The output of the \textit{listener} agent's language module, $g^l(\cdot)$, is the last hidden state of the LSTM layer, $h^l_L = g^l(m_L, h^l_{L-1})$.
In the context of the \textit{speaker} agent's language module $g^s(\cdot)$, the output is the message $m=(m_i)_{i\in[1,L]}$ consisting of one-hot encoded vectors of dimension $|V|$, which are sampled using the ST-GS approach from a categorical distribution $Cat(p_i)$ where $p_i = Softmax(\nu(h^s_i))$, provided $\nu$ is an affine transformation and $h^s_i=g^s(m_{i-1}, h^s_{i-1})$. $h^s_0=f(s_t)$ is the output of the visual module, given the target stimulus $s_t$.

\subsection{Decision Module}

Similar to \citet{Havrylov2017}, the decision module builds a probability distribution over a set of $K+1$ stimuli/images $(s_0, ..., s_K)$, consisting of the target stimulus and $K$ distractor stimuli, given a message $m$:
\[
    p((s_i)_{i\in[0,K]} | m) = Softmax( ( h^l_L \cdot f(s_i)^T )_{i\in[0,K]} ).
\]

\section{Lack of Correlation between Generalisation Abilities and Training-time Topographic Similarity}
\label{app:gen-f-ts-settings}

\begin{figure*}[b]
    \centering
    \includegraphics[width=1.0\linewidth]{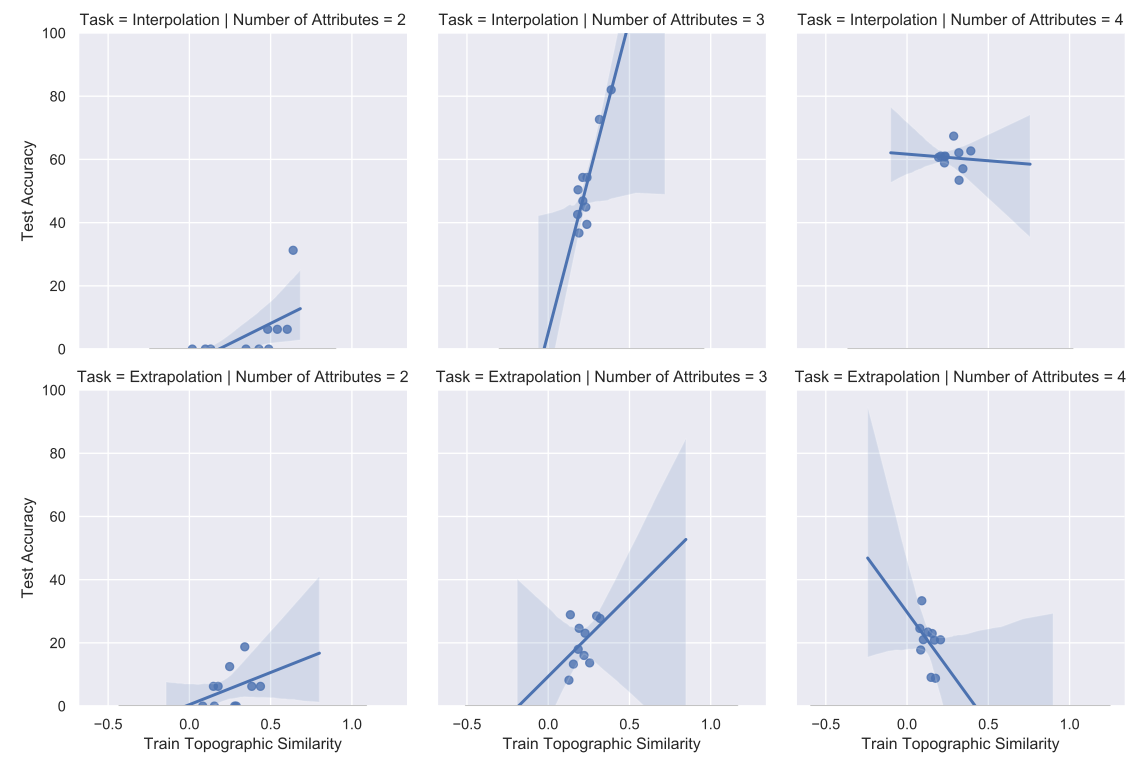}
    \caption{Training-time topographic similarity Vs. test-time accuracy  for each benchmark. A linear regression model is fit to the data for comparison.}
    \label{fig:toposim-struct-cap-breakdown}
\end{figure*}

Section~\ref{sec:exp:gen-compo-train} provided evidence that compositionality in the E-language at training-time, as reported by topographic similarity, cannot be expected to correlate with the generalisation abilities of the learning agents, without differentiating between the different settings. In this section, we provide a closer look at the results for each benchmark. Figure~\ref{fig:toposim-struct-cap-breakdown} shows the test-time accuracy with respect to the training-time topographic similarity for each different setting. The results of conducting Spearman rank-order correlation tests on the data yields correlation coefficients as described in Table~\ref{fig:spearman-rho-breakdown}. 

Strikingly, the results emphasise the difference between the interpolation task and the extrapolation task. While the results in the case of the interpolation task hint at a positive correlation hypothesis (the lower the level of structure in the observed meaning space is), those in the extrapolation task tend towards a negative correlation, for its most significant result (the higher the level of structure is). Our results are diverse and dependant on the level of structure in the observed meaning space, in addition to the type of task. This breakdown analysis further confirms our previous results that prevents us from rejecting the non-correlation hypothesis, and piles up further evidence to show that it would be incorrect to expect the training-time topographic similarity to predict any generalisation abilities of the learning agents.

\begin{table}[b]
    \centering
    \caption{Results of Spearman rank-order correlation tests for each benchmark, independent of the communication channel capacity (i.e. $5$ seeds in the complete case, and $5$ other seeds in the overcomplete case, for each benchmark).}
    \label{fig:spearman-rho-breakdown}
         \begin{tabular}{@{}l r r@{}} 
         \toprule
         Benchmark & Correlation Coefficient & P-value\\ 
         \midrule
         $T^2_{inter}$ & $0.811$ & $0.004$ \\ 
         $T^3_{inter}$ & $0.601$ & $0.066$ \\
         $T^4_{inter}$ & $0.115$ & $0.750$ \\
         \midrule 
         $T^2_{extra}$ & $0.356$ & $0.313$ \\ 
         $T^3_{extra}$ & $0.370$ & $0.294$ \\
         $T^4_{extra}$ & $-0.503$ & $0.138$ \\
         \bottomrule
        \end{tabular}
\end{table}

\section{Generalisation Difficulty in Interpolation \& Extrapolation}
\label{sec:gen-difficulty}

\textbf{Generalisation difficulty}, as defined by \citet{Chollet2019}, can be intuitively understood as a measure of how much the optimal behaviour of a skill program (here represented by the learning agent's weights) at evaluation-time differs from the optimal behaviour at training-time. The skill program is the product of an intelligent system (here represented by the whole optimisation process and surrounding algorithms that update the deep learning agents' weights), that we seek to evaluate the intelligence of. It is formally defined in an Algorithmic Information Theory fashion (considering program in terms of their description string), for a given task, $T$ (consisting of training and testing phases), a given curriculum, $C$ (here represented by the training scheme employed, i.e. a discriminative referential game), and a given skill threshold, $\theta$, as the ``length of the shortest program that, taking as input the shortest possible program that performs optimally over the situations in curriculum $C$, produces a program that performs at a skill level of at least $\theta$ during evaluation, normalized by the length of that skill program''~\citep{Chollet2019}. In other words, the (Relative) Algorithmic/Kolmogorov Complexity:
$GD^{\theta}_{T,C} = H(Sol^{\theta}_{T} | TrainSol^{opt}_{T,C} ) / H(Sol^{\theta}_{T})$
where $Sol^{\theta}$ is ``the shortest of all possible solutions to $T$ of threshold $\theta$ (shortest skill program that achieves at least skill $\theta$ during evaluation)'', and $TrainSol^{opt}_{T,C}$ is ``the shortest optimal training-time solution given a curriculum (shortest skill program that achieves optimal training-time performance over the [stimuli] in the curriculum)''. Therefore, $GD^{\theta}_{T,C}$ is the normalised description length (in a fixed universal language) of the shortest intelligent program that outputs $Sol^{\theta}_{T}$ when taking as input $TrainSol^{opt}_{T,C}$. 


\subsection{Experiments}
\label{sec:exp:gen-difficutly}

\begin{figure}[t]
    \centering
    \begin{subfigure}[t]{0.4\linewidth}
        \includegraphics[width=1.0\linewidth]{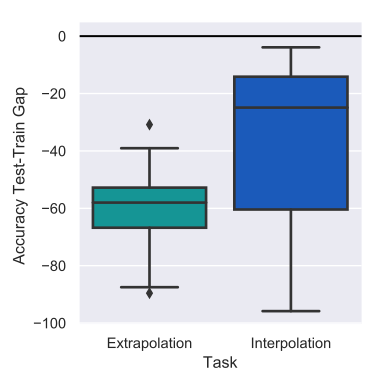}
    \end{subfigure}
    \begin{subfigure}[t]{0.4\linewidth}
        \includegraphics[width=1.0\linewidth]{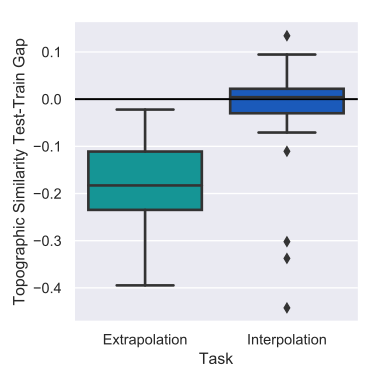}
    \end{subfigure}
    
    \caption{Distributions of the test-train gap in terms of accuracy and topographic similarity of emerging languages in different settings.
    }
    \label{fig:exp-ts-gen-gap}
    \label{fig:exp-acc-gen-gap}
\end{figure}


Following our primer on the definition of generalisation difficulty, in Section~\ref{sec:gen-difficulty}, we propose to investigate the generalisation difficulty of the interpolation task $T_{inter}$ and extrapolation task $T_{extra}$, for a skill threshold $\theta=38\%$ and our curriculum $C$ being a discriminative referential game with $K=47$ distractors and overcomplete communication channel. Defining optimality on the training phase of each task as achieving performance of at least $80\%$ accuracy, we find that all our different seeds in $T^3_{inter}$ and $T^3_{extra}$ can be considered optimal training-time solutions. We assume that any description of the learning agents are the shortest possible among the viable deep learning agents that can undertake our tasks. Our learning agents have minimal architecture design for language emergence and grounding (see Appendix~\ref{app:model-architecture} for details). It ensues that the identity function or program is a viable program to transform our distribution of shortest optimal training-time skill programs $TrainSol^{opt}_{T^3_{inter},C}$ into skill programs that perform at a skill level of $\theta=38\%$ on the 3-attributes interpolation task $T^3_{inter}$. We acknowledge that the identity program is the shortest description-length program possible on any fixed universal language, therefore it is the description-length of the shortest intelligent program that outputs $Sol^{\theta=38\%}_{T^{3}_{inter}}$ when taking as input $TrainSol^{opt}_{T^3_{inter},C}$. 

On the other hand, for the extrapolation task $T^{3}_{extra}$, Figure~\ref{fig:exp-acc-gen-gap}  shows that the identity program is not enough to transform the input $TrainSol^{opt}_{T^3_{extra},C}$ into $Sol^{\theta=38\%}_{T^{3}_{extra}}$, as the distributions of skill programs that are the shortest optimal training-time solutions on $T^{3}_{extra}$ perform below the skill level $\theta=38\%$. We interpret this result as the fact that the shortest description-length program transforming $TrainSol^{opt}_{T^3_{extra},C}$ into $Sol^{\theta=38\%}_{T^{3}_{extra}}$ must be more complex than the identity program (i.e. of a longer description-length). It ensues that $GD^{\theta=38\%}_{T^{3}_{extra},C} > GD^{\theta=38\%}_{T^{3}_{inter},C}$ (i.e. given the curriculum $C$ and skill threshold $\theta=38\%$, the generalisation difficulty of task $T^{3}_{extra}$ is higher than that of task $T^{3}_{inter}$). 

From a different and less formal viewpoint, Figure ~\ref{fig:exp-ts-gen-gap} shows the test-train gap in terms of topographic similarity of the E-language, across different settings. 
Independent of the level of structure in the meaning space, or the capacity of the communication channel, the topographic similarity test-train gap is consistently and significantly worse in the $T_{extra}$ 
($p\approx4e^{-10}$ for a one-sided Kolmogorov-Smirnov test with the alternative hypothesis being that the distribution of topographic similarity test-train gap in 
$T_{extra}$
is lower than the one of 
$T_{inter}$
), showing that the speaker agent fails to generalise systematically as the produced utterances at testing time are fairly different 
than at training time. We argue that this difference of behaviour that exists in 
$T_{extra}$ 
and not in 
$T_{inter}$ 
(the median of the distribution is at $0$), shows the extent with which the 
latter 
requires the behaviour of the agents to differ from the training time to the evaluation time.
As this is another intuition about generalisation difficulty~\citep{Chollet2019}, our results lend themselves to the conclusion that indeed there is some generalisation difficulty in 
$T_{extra}$ 
(in the order of at least $18\%$, as the median of the test-train gap in topographic similarity in $T_{extra}$), whilst there is none in 
$T_{inter}$. 

We expect this result to hold across curriculum for the family of visual referential games, independently of the technique that supports the communication channel, and for tasks where the stimuli are produced by a generative process, which take attribute vectors as input. For instance, referential games with symbolic stimuli are not expected to abide by these results.


\end{document}